\documentclass[runningheads]{llncs}
\usepackage[T1]{fontenc}
\usepackage{graphicx}
\usepackage{xcolor}
\usepackage[table]{xcolor}

\usepackage[linesnumbered,ruled,vlined]{algorithm2e}

\SetNlSty{nlstyle}{}{}

\usepackage{soul}
\usepackage{url}

\definecolor{hypercolor}{rgb}{0.0, 0.18, 0.39}

\usepackage{hyperref}
\hypersetup{
colorlinks=true,
linkcolor=blue!60!black,
citecolor=blue!60!black,
urlcolor=blue!60!black,
pdftitle={Research Proposal},
pdfauthor={Author Name},
pdfsubject={Research Proposal},
pdfcreator={XeLaTeX},
pdfproducer={LaTeX}
}

\usepackage[utf8]{inputenc}
\usepackage[T1]{fontenc}
\usepackage[small]{caption}
\usepackage{verbatim}
\usepackage{graphicx}
\usepackage{amsmath}
\usepackage{amssymb}
\usepackage{booktabs}
\usepackage[frozencache]{minted}
\usepackage{multirow}
\usepackage{float}
\usepackage{fontawesome5}
\usepackage{amsfonts}

\usepackage{subcaption}

\usepackage{minted}
\setminted{fontsize=\scriptsize}

\usepackage{lipsum}

\begin{document}

\title{Reasonable Motion}

\subtitle{{\normalsize A General ASP Foundation for Environment Constrained\\Movement Trajectory Computation}}
\titlerunning{\emph{Computing Reasonable Motion}}

\author{Julius Monsen\inst{1,3,} \and
Jakob Suchan\inst{2,3} \and
Mehul Bhatt\inst{1,3} \and
Lars Karlsson\inst{1,3}}

\authorrunning{Monsen et al.}

\institute{Örebro University, Örebro, Sweden\\
\email{info@codesign-lab.org} \and
Constructor University Bremen, Bremen, Germany\\
\email{jsuchan@constructor.university} \and
CoDesign Lab > Cognitive Vision\\
\href{https://codesign-lab.org/cognitive-vision/}{codesign-lab.org/cognitive-vision}}

\maketitle

\begin{abstract}
We present a general \emph{answer set programming} based hybrid quantitative-qualitative method for computing constrained branching trajectory modes for moving objects in real-world settings. The method performs constrained traversal of an environment graph, enumerating geometrically admissible motion behaviours as stable models, each constituting a distinct trajectory mode characterised by both domain-dependent and independent factors such as derived event sequence, map topology, and domain norms. The hybrid trajectory computation method is generally applicable across motion characteristics typically encountered in diverse dynamic domains with moving objects, e.g., autonomous driving. We demonstrate applicability and highlight how computed trajectories are traceable to their underlying stable model, thereby affording verifiable interpretability that purely learned approaches cannot provide. We also perform an empirical evaluation with Argoverse~2, a large-scale real-world autonomous driving benchmark representative of the class of dynamic domains within the scope of the proposed method.
\end{abstract}

\section{Motivation}
Autonomous systems operating in embodied \textbf{in-the-wild} situations are confronted with the challenge of \emph{interactive sensemaking}, i.e., interleaved multimodal perception, interpretation, and decision-making requiring coordination of attention, integration of sensory inputs, and dynamic exploration of
possible futures and interactive outcomes \cite{Bhatt-AVI,kr2025-asp-visualcomm}. Such capabilities, referred to as \textbf{physical~AI}, are gaining renewed traction as a major challenge for contemporary AI/ML research, and will be critical for deploying trustworthy technologies in safety-critical dynamic domains such as social robotics and autonomous driving.

\smallskip \textbf{Motion in Structured Environments.}\quad
Many real-world domains in which autonomous agents operate are structured: roads constrain vehicle motion to lanes and require normative compliance in intersections and roundabouts; indoor environments constrain robot navigation to corridors, doorways, and rooms; social spaces impose normative constraints on pedestrian flow. This structure is not merely a backdrop; it is a rich source of information about what motions are geometrically admissible, physically plausible, and normatively appropriate. For instance, a human driver approaching an intersection does not reason over a continuous positional space; rather, they reason over a small set of symbolically distinct possibilities (\emph{go straight}, \emph{turn left}, \emph{change lane}) that are immediately grounded in the geometry of the road ahead. Capturing this kind of {geometry-constrained, commonsense qualitative inference} \cite{Bhatt-AVI,DBLP:journals/jair/Davis17,DBLP:journals/cacm/DavisM15} in a principled, transparent, and formally verifiable manner remains an open challenge, particularly in view of emerging AI regulation and the need for retrospective diagnosability in safety-critical deployments \cite{Regulation-2021}.

\smallskip \textbf{Limits of End-to-End Motion Prediction.}\quad
Purely end-to-end stochastic approaches for the prediction of motion trajectories (of dynamic interacting objects) often focus on learning functional mappings from scene observations to future positions~\cite{shi2025motionsurvey}, without explicit representations of the environmental structure, admissible behaviours, or the events that characterise them. Such end-to-end approaches emphasise purely geometric accuracy on benchmarks, prioritising \emph{where} agents may move, offering little insight into \emph{why} a particular motion trajectory was generated or prioritised, or \emph{which} aspects of the environment constrained it~\cite{shi2025motionsurvey,zhou2023query,zhang2024demo,knoche2025donut}. Qualitatively distinct behaviours (e.g., going straight versus turning at an intersection) may be represented as overlapping geometric trajectories, and there might not be a systematic semantic mechanism by which such distinctions can be inspected, explained, verified, or evaluated at scale.

\smallskip \textbf{Key Contributions.}\quad
We present an answer set programming (ASP) \cite{ASP-Glance-2011,Gebser2014-Clingo} based \emph{hybrid quantitative-qualitative} method for computing `reasonable' motion trajectory modes for moving objects in real-world dynamic settings, that are geometrically admissible and normatively preferred with respect to the structured environment representation and domain constraints. The developed {hybrid} method consists of: {\small\bfseries 1)}~an ASP encoding qualitatively modelling the environment as a typed directed graph, and performing constrained traversal to systematically enumerate geometrically admissible, branching motion trajectory modes; and {\small\bfseries 2)}~a trajectory generation component that converts each qualitatively distinct motion mode into a smooth, continuous motion trajectory via environmental geometry-guided centerline construction and constant-speed following.

\smallskip

Our modular hybrid qualitative-quantitative characterisation ensures separation of concerns between the high-level commonsense and low-level numerical levels of representation of motion. Furthermore, we emphasise that following our approach, each computed motion trajectory is characterised by a verifiable path and a high-level event sequence capturing domain-dependent and independent factors such as map topology, norms/preferences. To highlight real-world relevance, we empirically evaluate the method in the autonomous driving domain using the Argoverse~2 community benchmark \cite{Argoverse2}. Here, we also demonstrate how every computed motion trajectory is fully traceable to its underlying stable model, thereby affording verifiable interpretability that most purely learned approaches lack. We posit that the proposed method is a principled---i.e., domain-independent, modular, and elaboration tolerant---and transparent alternative to monolithic end-to-end predictors in domains where explainability and geometric admissibility are critical requirements.

\section{Computing `Reasonable' Motion: A Hybrid Qualitative-Quantitative Method}
\label{sec:formal-framework}
The proposed \emph{hybrid qualitative-quantitative method} for computing (reasonable) motion trajectories is illustrated in Fig.~\ref{fig:system-diagram}. Operationally, the method consists of four stages from a higher-level viewpoint: {\small\bfseries {(1)}} Grounding the \emph{answer set program}\footnote{Answer Set Programming (ASP) is an established declarative programming paradigm rooted in the stable model semantics; we refer readers to \cite{ASP-Glance-2011,Gebser2014-Clingo,ASP-books/sp/Lifschitz19} for an introduction and further pointers.} from the scene description and deriving reachability over the environment graph; {\small\bfseries {(2)}} Path selection via a choice rule and integrity constraints; {\small\bfseries {(3)}} Grouping the resulting stable models by event sequence; and {\small\bfseries {(4)}} Converting each mode group into a continuous trajectory. \emph{A formal characterisation of the method follows}:

\begin{figure}[t]
    \centering
    \includegraphics[width=1.0\linewidth]{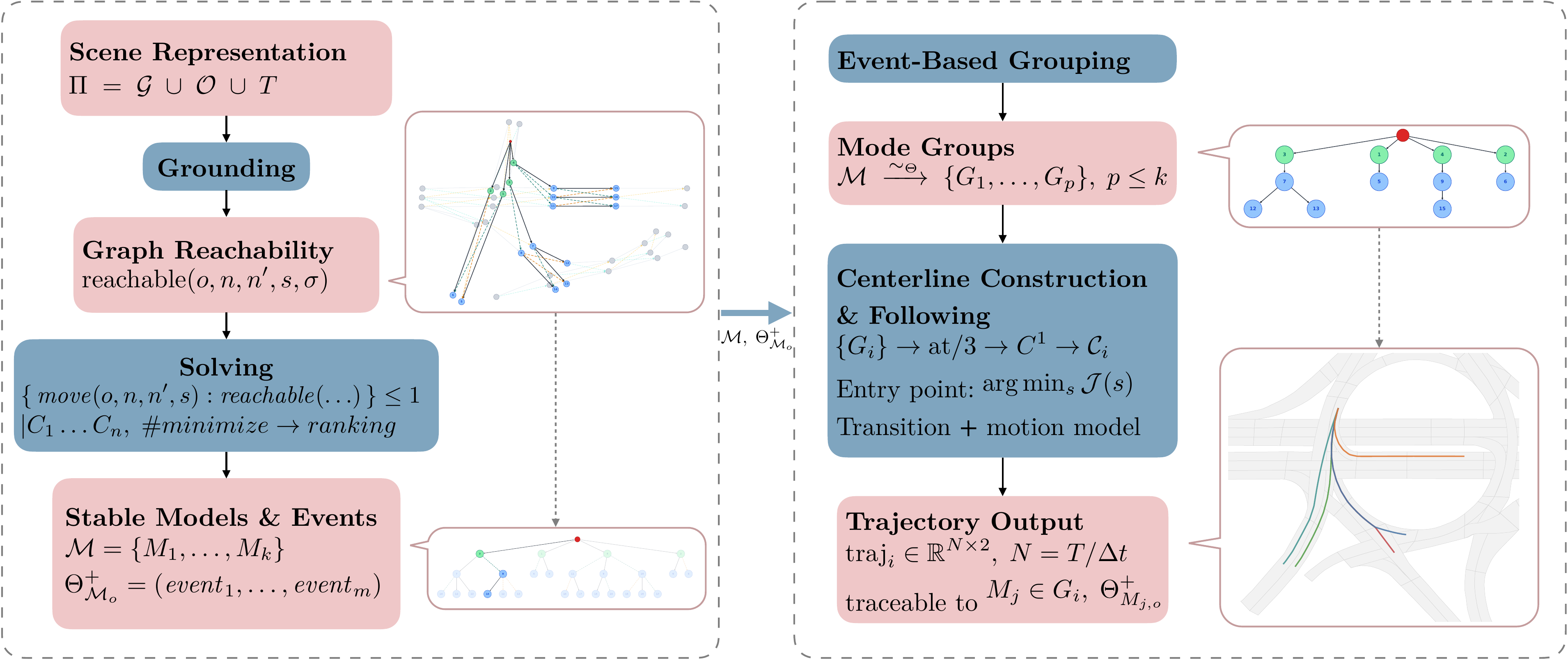}
    \caption{{\bfseries Hybrid Qualitative-Quantitative Motion Trajectory Computing Method}: Overview of the proposed pipeline. \textbf{Left:} ASP grounding and solving yield stable models, each representing a path annotated trajectory influencing events.  \textbf{Right:} stable models are grouped and geometrical features constructed such that paths can be followed continuously according to a domain-specific motion model and yield a trajectory $\mathrm{traj}_i \in \mathbb{R}^{N \times 2}$ per mode group.}
    \label{fig:system-diagram}
\end{figure}

\subsection{Scene Representation}
\label{sec:modelling}
Let $\mathcal{O} = \{o_1, \ldots, o_n\}$ be a set of domain objects (e.g., moving entities), each associated with a state $o^{\mathrm{state}}$  comprising position, velocity, and heading at a given timestep. The environment is characterised by a finite set of discrete spatial regions $\mathcal{N}$, a set of time steps $\mathcal{T} = \{t_0, \ldots, t_m\}$, and a prediction horizon $T \subset \mathcal{T}$ over which trajectory modes are computed. The ASP program encoding the scene is then: $\Pi \;=\; \mathcal{G} \;\cup\; \mathcal{O} \;\cup\; T$, where $\mathcal{G} = (\mathcal{N}, \mathcal{E})$ is a typed directed graph whose nodes and edges are grounded as facts, while $\mathcal{O}$ and $T$ are collections of facts encoding object states and time steps, respectively. Nodes $n \in \mathcal{N}$ represent discrete spatial regions and edges $(n_1, n_2, \delta) \in \mathcal{E}$ carry typed relations $\delta \in \Delta$ encoding navigational affordances (e.g., \emph{successor} for sequential region transitions, \emph{lateral} for side-by-side transitions such as lane changes, and \emph{entry} for access relations into junctions or restricted zones). Each node is annotated with domain-dependent attributes such as region type (e.g., vehicle lane, bike lane, intersection), geometric extent (e.g., arc length or bounding polygon), or normative constraints (e.g., permitted crossing directions encoded in lane markings). A \emph{trajectory mode} for object $o$ is characterised by a derived event sequence $\Theta_o^{+}$ capturing qualitatively distinct behavioural events, grounded in a supporting node sequence $(n_0, n_1, \ldots, n_k)$ through $\mathcal{G}$, constituting a commonsense / qualitative hypothesis about the future behaviour of $o$.

\begin{figure}[t]
    \centering
    \includegraphics[width=\linewidth]{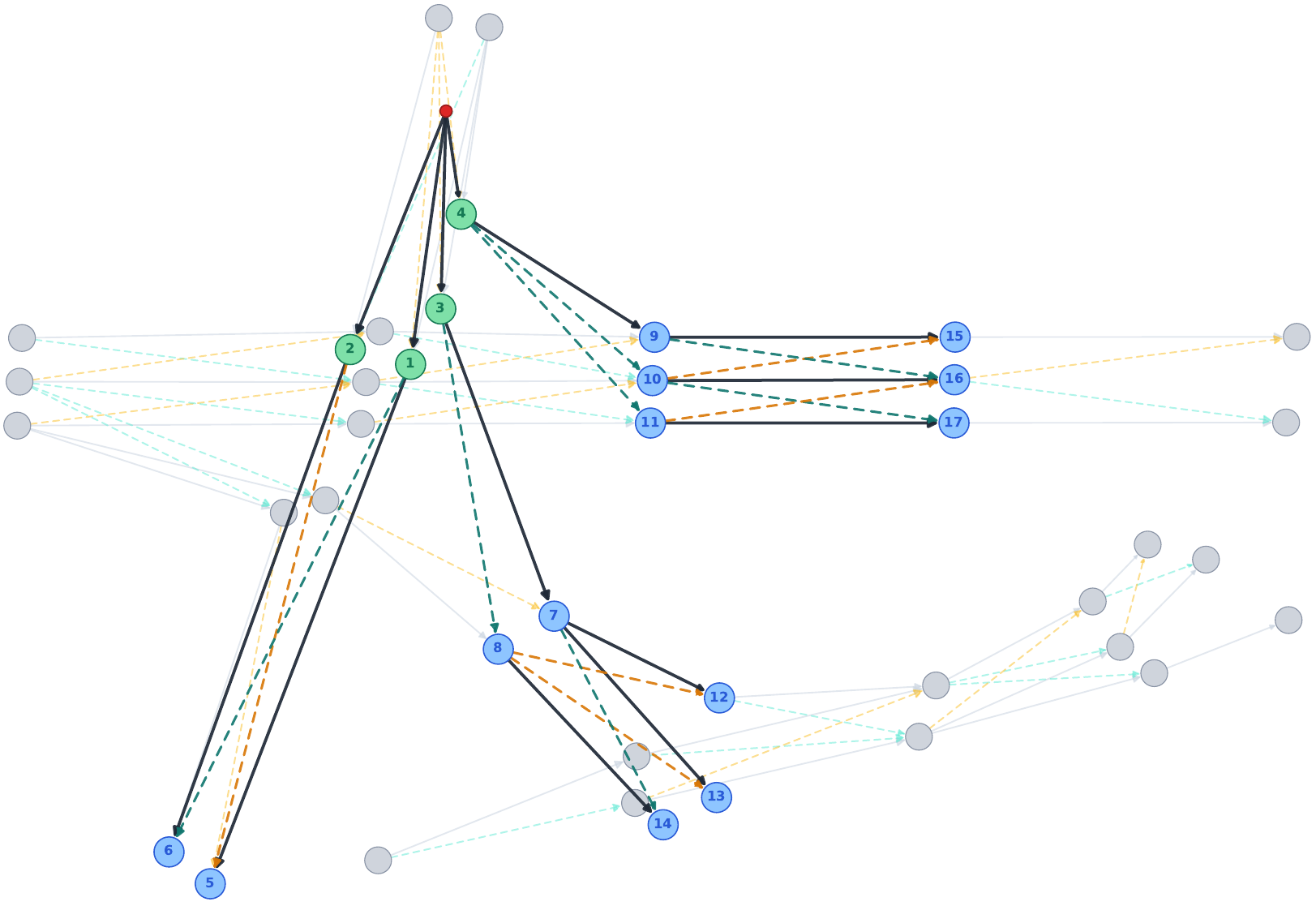}
    \caption{{\bfseries Reachability} over $\mathcal{G}$; Example:
    \textcolor[HTML]{dc2626}{$\bullet$}~\emph{current agent position};
    \textcolor[HTML]{86efac}{$\bullet$}~\emph{potential start node};
    \textcolor[HTML]{93c5fd}{$\bullet$}~\emph{reachable nodes}. Solid arrows
    denote successor transitions; dashed arrows denote two-hop lateral
    relations (as used in Sec.~\ref{sec:asp-mode-gen}).}
    \label{fig:abstract-graph}
\end{figure}

\subsection{Graph Reachability}
\label{sec:reachability}
Reachability over $\mathcal{G}$ is computed as a derived relation that tracks, for each object $o$, the set of nodes reachable within the prediction horizon, together with a \emph{traversal state} $\sigma$ recording accumulated progress and the history of any mode-changing transitions relevant to the domain (e.g., distance travelled and lateral manoeuvre history in driving, or floor level and door transitions in indoor navigation): $\mathrm{reachable}(o,\, n,\, n',\, s,\, \sigma)$. Rather than embedding recursive traversal directly inside the choice rule, reachability is defined separately as an inductively derived relation over the environment graph. This follows a modelling pattern used for recursively defined concepts such as graph reachability~\cite{ASP-Glance-2011}, where auxiliary Horn rules derive the transitive structure independently of the non-deterministic path selection. Operationally, this yields a relatively flat search problem for the solver. Rather than resolving recursive path dependencies during search, the solver operates over a pre-instantiated set of independent ground transitions, reducing backtracking and conflict depth. The traversal state $\sigma$ and its propagation rules remain domain-dependent. An illustration of reachability in a concrete graph is shown in Fig.~\ref{fig:abstract-graph}, where an agent is positioned at the red marker.

\subsection{Constrained Mode Hypothesis Generation}
\label{sec:path-selection}
Mode hypotheses are generated by selecting a path through the reachable set via an ASP choice rule:
{\small
\begin{equation}
  \bigl\{\;\mathrm{move}(o, n, n', s) :
    \mathrm{reachable}(o, n, n', s, \sigma)\;\bigr\} \leq 1
  \qquad \forall\, s \in \mathrm{steps}(o)
\end{equation}
}

The set of stable models of $\Pi$ under this rule is denoted $\mathcal{M} = \{M_1, \ldots, M_k\}$, each constituting a distinct, internally consistent trajectory mode hypothesis. Path validity (physical and normative) is enforced by the following integrity constraints, which hold across all domain instantiations:

\smallskip

{\small\bfseries C1}. {\bfseries Mandatory start}: $\leftarrow \neg\,\mathrm{move}(o, \cdot, \cdot, 0)$\\[2pt]
    Every object must be assigned a transition at step $s=0$,
    grounding the trajectory in the object's observed state at the
    first prediction timestep.
    
\smallskip

{\small\bfseries C2}. {\bfseries Temporal consistency}:
      $\leftarrow \mathrm{move}(o, \cdot, n', s),\;
    \mathrm{move}(o, n, \cdot, s{+}1),\; n \neq n'$. \\[2pt]
    The destination node at step $s$ must equal the origin node at
    step $s{+}1$, enforcing that selected transitions form a connected
    path through $\mathcal{G}$.

    \smallskip

{\small\bfseries C3}. {\bfseries Continuity}:    
    $\leftarrow \neg\,\mathrm{move}(o, \cdot, \cdot, s),\;
    \mathrm{move}(o, \cdot, \cdot, s{+}1)$. \\[2pt]
    A transition at step $s{+}1$ is only permitted if a transition was
    chosen at step $s$. Together with C2, this ensures the selected
    path is contiguous: once a path terminates, no further transitions
    are admitted.

\smallskip

Additional domain-dependent constraints, e.g., bounding the number and spacing of mode-changing transitions, can be incorporated per domain. Stable models may furthermore be ranked via ASP weak constraints encoding domain-dependent preferences $\mathcal{W}$---such as penalising lane changes across solid lane markings or favouring paths requiring fewer lateral transitions---yielding an ordered set of trajectory mode hypotheses from which downstream components can select or sample. Each $\mathrm{move}(o, n, n', s)$ atom additionally derives $\mathrm{at}(o, n', s)$, recording the node occupied by $o$ at step $s$, used for event derivation in Sec.~\ref{sec:grouping}. Exemplary ASP implementation of both domain-specific constraints and preference ranking is included in Sec.~\ref{sec:asp-mode-gen}.

\subsection{Event-Based Grouping}\label{sec:grouping}
Following solving, the resulting stable models are grouped as a post-processing step based on their derived event sequences. To provide interpretable characterisations of each mode and enable grouping of behaviourally equivalent stable models, high-level events are derived directly from the selected path:
{\small
\begin{equation}
  \mathrm{event}(e(o, \ldots), s) \leftarrow
    \mathrm{at}(o, n, s),\; \phi(o, n, s)
\end{equation}
}

where $\phi(o, n, s)$ is a domain-supplied condition on the current node and step. The ordered tuple of event atoms derived from each stable model $M_i \in \mathcal{M}$ serves as a qualitative (event) fingerprint $\Theta_{\mathcal{M}_o}^{+} = (\mathit{event}_1, \ldots, \mathit{event}_m)$, 
ordered by step index $s$ so that simultaneous events at the same step are canonically sorted. Models sharing the same fingerprint, i.e., $M_i \sim_\Theta M_j \;\iff\; \Theta_{M_i}^{+} = \Theta_{M_j}^{+}$, are collapsed into a single mode group $G_i$ via the equivalence relation $\sim_\Theta$:
{\small
\begin{equation}
  \mathcal{M} \;\xrightarrow{\;\sim_\Theta\;}\;
  \{G_1, \ldots, G_p\}, \quad p \leq k
\end{equation}
}

This grouping serves two purposes: {\small\bfseries(1)} reduction of the mode space to a manageable set of behaviourally distinct hypotheses (the number of stable models can be large in complex scenarios with many reachable nodes and permitted transitions); and {\small\bfseries(2)} enabling an interpretable high-level description of each mode (e.g., \emph{lane change left, go straight}, or \emph{turn right at intersection}) that can be inspected for transparency and used to guide downstream trajectory generation together with the resulting lane occupation information.

\subsection{From Commonsense Qualitative Modes to Continuous Geometric Trajectories}
\label{sec:traj-gen-general}
Given the grouped stable models, each mode group $G_i$ is converted into a single continuous trajectory in two steps.
 
{\small\bfseries(1). Geometric Mode Construction}.\quad Each stable model $M_j \in G_i$ yields an ordered sequence of nodes, from which geometric representations (e.g., centerlines or region boundaries) are retrieved. Within a group, different models may follow geometrically distinct but behaviourally equivalent paths. To preserve the qualitative behaviour while reducing mode redundancy, a single representative, approximate geometric path $\mathcal{C}_i$ is obtained by stitching and averaging the node geometries across all models in $G_i$, ensuring $C^1$-continuity at junctions. Since all models in $G_i$ share the same event fingerprint $\Theta^+_{G_i}$ by construction, the averaged path preserves the qualitative behavioural identity of the group while geometrically centralising across lane-level variations within it. For instance, if two models both perform a left lane change but at adjacent lane steps, the averaged path will execute the change at an intermediate position. The specific stitching and averaging procedure is domain-dependent and described in Sec.~\ref{sec:trajectory-generation}.

{\small\bfseries(2). Trajectory Following}.\quad Given $\mathcal{C}_i$ and the object's observed state $o^{\mathrm{state}}$, a trajectory is produced over the planning horizon $T$. If the object is not already on $\mathcal{C}_i$, a smooth transition curve connects
the current position and heading to an entry point on $\mathcal{C}_i$. The entry point is selected by minimising an angular cost penalising misalignment between the object's heading and the straight line to the
entry point, and between that line and the path tangent at entry, with a distance regularisation term:
{\small
\begin{equation}
  \mathcal{J}(s) = (1 - \cos\theta_{\mathrm{depart}})
                 + (1 - \cos\theta_{\mathrm{arrive}})
                 + \lambda\,\frac{\|p(s) - p_0\|}{L},
  \label{eq:entry-cost}
\end{equation}
}
where $\theta_{\mathrm{depart}}$ is the angle between the object's heading and the line to entry point $p(s)$, $\theta_{\mathrm{arrive}}$ is the angle between that line and the path tangent at $p(s)$, $L$ is the total path length, and $\lambda$ is a distance weight. The object then follows $\mathcal{C}_i$ from the entry point under a chosen motion model for the remainder of $T$, yielding:
{\small
\begin{equation}
  \mathrm{traj}_i \in \mathbb{R}^{N \times 2}, \quad N = T / \Delta t
\end{equation}
}
Each trajectory $\mathrm{traj}_i$ is fully traceable to its underlying stable models $\{M_j \in G_i\}$ and event sequence $\Theta_o^{+}$, affording verifiable interpretability over the complete pipeline. In Alg.~\ref{alg:mode_computation} we highlight the flow of the overall system and each functional component as outlined in this section. 

\begin{algorithm}[t]
\caption{Hybrid Trajectory Mode Computation}
\label{alg:mode_computation}
\footnotesize

\KwIn{$\Pi = \mathcal{G} \cup \mathcal{O} \cup T$,
      constraints $\{C_1,\ldots,C_n\}$,\\preferences $\mathcal{W}$, agent state $o^{\mathrm{state}}$, horizon $T$, $\Delta t$}
\KwOut{$\mathcal{Y} = \{(\mathrm{traj}_i,\,\Theta_{G_i}^{+})\}_{i=1}^{p}$}

$\mathcal{R} \leftarrow \mathrm{ground}(\Pi)$ \tcp{derive all $\mathrm{reachable}(o,n_{\mathrm{prev}},n,s,\sigma)$ at ground time}

{\color{hypercolor}\tcp*[h]{Mode Hypothesis Generation}}

$P_{\mathrm{move}}
=
\Bigl\{
\{
\mathrm{move}(o,n,n',s)
:
\mathrm{reachable}(o,n,n',s,\sigma)
\}1
\Bigr\}$

$P \leftarrow
\mathcal{R}
\cup
P_{\mathrm{move}}
\cup
\{C_1,\ldots,C_n\}
\cup
\mathcal{W}$;\quad $\mathcal{M} \leftarrow \mathrm{optAnswerSets}(P)$

{\color{hypercolor}\tcp*[h]{Event-Based Grouping}}

$\mathcal{F} \leftarrow \{\}$ 

\ForEach{$M_i \in \mathcal{M}$}{
  $E_i \leftarrow \emptyset$
  
  \ForEach{$(o, n, s)$ s.t.\ $\mathrm{at}(o,n,s) \in M_i$}{
    \If{$\phi(o, n, s)$ holds}{
      $E_i.\mathrm{add}\bigl(\,(e(o,\ldots),\, s)\,\bigr)$
    }
  }
  $\Theta_{M_i}^{+} \leftarrow \mathrm{sort}(E_i)$;\quad $\mathcal{F}[\Theta_{M_i}^{+}].\mathrm{add}(M_i)$
}
$\{G_1,\ldots,G_p\} \leftarrow \mathrm{values}(\mathcal{F})$

{\color{hypercolor}\tcp*[h]{Trajectory generation}}\\
$\mathcal{Y}  \leftarrow \emptyset$

\ForEach{$G_i$}{
  $\mathcal{P} \leftarrow \emptyset$
  
  \ForEach{$M_j \in G_i$}{
    $n_1,\ldots,n_k \leftarrow \mathrm{at}$-sequence of $M_j$
    
    $\mathcal{P}.\mathrm{add}\bigl(\mathrm{stitch}(n_1,\ldots,n_k,\,\mathcal{G})\bigr)$
  }
  
  $\mathcal{C}_i \leftarrow \mathrm{arc\_length\_average}(\mathcal{P})$;\quad  $s^* \leftarrow \arg\min_{s}\;\mathcal{J}(s;\;o^{\mathrm{state}},\mathcal{C}_i)$
  
  $\mathrm{traj}_i \leftarrow \mathrm{follow}(\mathcal{C}_i,\, s^*,\, o^{\mathrm{state}},\, N{=}T/\Delta t)$;\quad    $\mathcal{Y}.\mathrm{add}\bigl((\mathrm{traj}_i,\,\Theta_{G_i}^{+})\bigr)$
}
\Return $\mathcal{Y}$
\end{algorithm}

\section{Reasonable Motion in Autonomous Driving}\label{sec:implementation}

We apply and practically demonstrate an encoding of the proposed (reasonable) motion trajectory computation method (Sec.~\ref{sec:formal-framework}) for the domain of autonomous driving, where a typical requirement\footnote{Section \ref{sec:eval} includes an empirical evaluation with the \emph{in-the-wild} Argoverse~2 \cite{Argoverse2} motion forecasting community benchmark consisting of complex urban driving scenarios together with corresponding motion data from LIDAR and RGB signals. Here, the motion prediction task operates over a 6-second horizon at 10\,Hz.}  is to compute motion trajectories of dynamic entities. By \emph{reasonable} motion, we allude to computed solutions that are both formally well-founded and behaviourally well-justified. Here, two connotations are implied:  {\small\bfseries(1)} in the manner of non-monotonic commonsense reasoning, our model derives motion trajectories that are ``reasoned about'': each motion hypothesis is grounded in a stable model, fully traceable to its supporting ASP derivation, and thereby is amenable to inspection and explanation; and {\small\bfseries(2)} we also allude to ``reasonableness'' as a solution criterion, i.e., only trajectories satisfying explicit geometric, physical, and contextual or norm-respecting constraints---encoded as integrity constraints and ranked by preference minimisation---are admitted.

\subsection{High-Level Commonsense Mode Hypothesis Generation}\label{sec:asp-mode-gen}
Given an ASP problem specification for a motion trajectory computation task in the domain of driving, it is desired that each stable model correspond to a distinct trajectory mode---a high-level behavioural hypothesis capturing the sequence of lanes a vehicle may occupy and the discrete events arising along it (e.g., \emph{go straight}, \emph{turn left}, \emph{change lane then go straight})---guiding downstream trajectory generation (Sec.~\ref{sec:trajectory-generation}).

\subsubsection{Lane Graph}
The environment graph $\mathcal{G}$ is instantiated with lane segments $\mathcal{L} = \{\ell_1, \ldots, \ell_n\}$ as nodes. Each edge $(\ell_1, \ell_2, \delta) \in \mathcal{E}$ carries a relation type $\delta \in \{\mathrm{successor}, \mathrm{left}, \mathrm{right}\}$, and each lane $\ell$ is annotated with its arc length $\mathrm{dist}(\ell) \in \mathbb{R}^{+}$, its type (vehicle, bus, bike), and its lane markings on each side. A subset $\mathcal{L}_I \subseteq \mathcal{L}$ of intersection lanes are tagged with a turn direction $\mathrm{dir}_I(\ell) \in \{\mathrm{left}, \mathrm{right}, \mathrm{straight}\}$.

Lane changes, the domain instantiation of mode-changing transitions, are derived as two-hop paths in $\mathcal{G}$ (Eqs.~\eqref{eq:lc1}--\eqref{eq:lc2}). A third case (Eq.~\eqref{eq:lc3}) captures the common manoeuvre where a turning vehicle drifts into an adjacent parallel exit lane upon leaving an intersection. Let $\ell_1$ be the turning lane with direction $\tau$, $m$ its primary exit, and $\ell_2$ the adjacent lane in opposite direction $\bar\tau$ relative to the turn direction. Then a connection to $\ell_2$'s continuation $\ell_3$ is admitted.

{\small 
\begin{align}
  \mathrm{lc}(\ell_1, \ell_2, \delta) &\leftarrow
    (\ell_1, m, \delta)\!\in\!\mathcal{E},\;
    (m, \ell_2, \mathrm{succ})\!\in\!\mathcal{E},\;
    \delta \neq \mathrm{succ} \label{eq:lc1}\\
  \mathrm{lc}(\ell_1, \ell_2, \delta) &\leftarrow
    (\ell_1, m, \mathrm{succ})\!\in\!\mathcal{E},\;
    (m, \ell_2, \delta)\!\in\!\mathcal{E},\;
    \delta \neq \mathrm{succ} \label{eq:lc2}\\
  \mathrm{lc}(\ell_1, \ell_3, \bar\tau) &\leftarrow
    (\ell_1, m, \mathrm{succ})\!\in\!\mathcal{E},\;
    \ell_1 \in \mathcal{L}_I,\;
    \mathrm{dir}_I(\ell_1) = \tau, \notag \\
  &\phantom{{}\leftarrow{}}
    (m, \ell_2, \bar\tau)\!\in\!\mathcal{E},\;
    (\ell_2, \ell_3, \bar\tau)\!\in\!\mathcal{E}
    \label{eq:lc3}
\end{align}
}

\subsubsection{Traversal State and Reachability}
The general traversal state $\sigma$ (see Sec.~\ref{sec:reachability}) is instantiated for driving as $\sigma = (d, \delta, \kappa, d_\kappa, \delta_\kappa)$, where $d$ is the accumulated distance, $\delta$ the edge type of the transition from $\ell_{\mathrm{prev}}$ to $\ell$, $\kappa$ the lane change count, $d_\kappa$ the cumulative distance at the most recent lane change, and $\delta_\kappa$ its direction. This yields the following predicate schema:

\begin{minted}[fontsize=\scriptsize,bgcolor=blue!5!white]{prolog}
reachable(Trk, L1, L2, Step, sigma(D, Delta, K, DK, DeltaK)) :- ...
\end{minted}
 
The step index orders lane visits but does not correspond to a fixed time interval (lanes vary in length). Lane-vehicle type compatibility is enforced via \texttt{allowed\_in\_lane/2}, and the recursion is bounded by $d_{\mathrm{max}}$ derived from the observed speed under a chosen motion model.

\paragraph{Start.}
Each vehicle is seeded at its potential start lane(s) with $s=0$, $\kappa=0$, $\delta_\kappa=\mathrm{none}$. Occupied lane(s) and remaining distance $d_0$ are determined by querying spatially nearby lanes, scoring candidates by heading alignment and containment, and applying robustness heuristics for sensor noise and low-speed ambiguity. If the vehicle is sufficiently far from the lane end, neighbouring lanes of adequate length are also admitted as start lanes.

\paragraph{Successor propagation.}
From any state at step $s$ with $d_{\mathrm{old}} < d_{\mathrm{max}}$, the vehicle advances to a successor lane $\ell'$, accumulating $d = d_{\mathrm{old}} + \mathrm{dist}(\ell')$ while the lane-change state is carried forward unchanged.

\paragraph{Lane change propagation.}
A lateral move via $\mathrm{lc}(\ell, \ell', \delta)$ is permitted provided $\kappa < \kappa_{\max}$ and $\ell'$ is type-compatible. Three cases govern direction and state update, all incrementing $\kappa$ and recording $d_{\mathrm{old}}$ as the new $d_\kappa$: (i)~first change ($\delta_\kappa = \mathrm{none}$): any direction permitted, $\delta_\kappa$ set to $\delta$; (ii)~continued direction ($\delta = \delta_\kappa$): always permitted, $\delta_\kappa$ unchanged; (iii)~opposite direction ($\delta = \bar\delta_\kappa$): permitted only if a minimum recovery distance has been observed since the last lane change $d_{\mathrm{old}} - d_\kappa \geq \theta_{\mathrm{lc}}$, then $\delta_\kappa$ updated to $\delta$.

\subsubsection{Mode Selection, Ranking and Events} 
Applied to the driving domain, the mode selection choice rule (Sec.~\ref{sec:path-selection}) takes the form:
\begin{minted}[fontsize=\scriptsize,bgcolor=blue!5!white]{prolog}
{ move(Trk, L1, L2, S) : reachable(Trk, L1, L2, S, _) } 1 :- step(Trk, S).
\end{minted}

In addition to the general constraints C1--C3 described in Sec.~\ref{sec:path-selection}:
\begin{minted}[fontsize=\scriptsize,bgcolor=blue!5!white]{prolog}
:- not move(Trk, _, _, 0), step(Trk, 0).                                               %(C1)
:- move(Trk, _, L2, S), move(Trk, L1, _, S+1), L1 != L2.                               %(C2)
:- step(Trk, S), not move(Trk, _, _, S), move(Trk, _, _, S+1).                         %(C3)
\end{minted}

the following driving-specific constraints are enforced: (C4)~minimum travel of at least the braking
distance; (C5)~lane change budget $\kappa_{\max}$; and (C6)~minimum spacing $\theta_{\mathrm{lc}}$ between opposite-direction changes:

\begin{minted}[fontsize=\scriptsize,bgcolor=blue!5!white]{prolog}
:- move(Trk, L1, L2, S), not move(Trk, _, _, S+1), braking_distance(Trk, BD), D < BD, 
   reachable(Trk, L2, _, _, _), reachable(Trk, L1, L2, S, sigma(D, _, _, _, _)).       %(C4)
:- #count{ S : changed_lane(Trk, S, _) } > max_lane_changes.                           %(C5)
:- consecutive_lc(Trk, S1, S2), #sum{ D, S : move(Trk, _, L2, S), 
   lane_distance(L2, D), S>=S1, S<S2}  < lc_distance_threshold.                        %(C6)
\end{minted}

\texttt{braking\_distance/2} is provided as a fact from the observed speed. The predicates \texttt{changed\_lane/3} and \texttt{consecutive\_lc/3} are derived helpers capturing lateral transitions and consecutive opposite-direction pairs respectively. Each \texttt{move/4} is converted to \texttt{at/3} representing lane occupation.

Stable models are ranked by a two-level minimisation encoding traffic norm compliance via the map's \texttt{lane\_mark/3} facts. At higher priority, lane changes across transitional markings (e.g.,
\texttt{dash\_solid}) incur cost~1 and solid or double-solid markings incur cost~5. At lower priority, total lane change count is minimised and better start-lane heading alignment is preferred. The solver is configured with \texttt{--opt-mode=optN} and \texttt{models=0}, directing Clingo to enumerate \emph{all} cost-optimal stable models. This ensures that every behaviourally distinct trajectory mode---not only the single highest-ranked one---is retained for downstream grouping and trajectory generation.

\begin{minted}[fontsize=\scriptsize,bgcolor=blue!5!white]{prolog}
#minimize { C@2, Trk, S, Dir : changed_lane(Trk, S, Dir),
            move(Trk, L1, _, S), lc_mark_cost(L1, Dir, C) }.
#minimize { 1@1, S, Trk : changed_lane(Trk, S, _) }.
#minimize { Cost@1, Trk : in_lane(Trk, L, curr_time, D, Cost) }.
\end{minted}

\begin{figure}[t]
    \centering
    \includegraphics[width=0.9\linewidth]{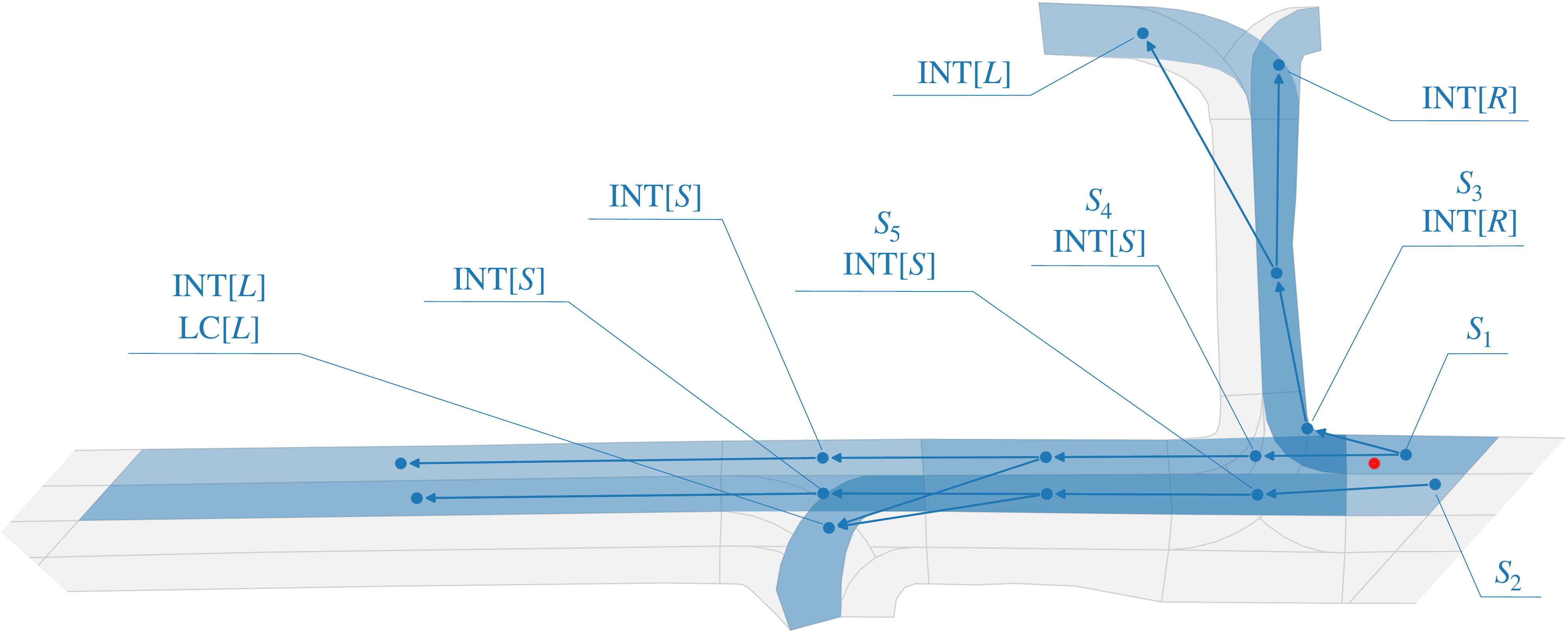}
    \caption{{\bfseries Discrete Trajectory Modes}. Each mode is highlighted by a path between the lane segments and their representative points \textcolor[HTML]{1f77b4}{$\bullet$}. \textcolor[HTML]{dc2626}{$\bullet$} denotes the position of the agent. {\sffamily INT[S/L/R]} denotes an intersection event where the agent goes \emph{straight}, \emph{left} or \emph{right}, and  {\sffamily LC[L/R]} a \emph{left} or \emph{right} lane change.}
    \label{fig:modes}
\end{figure}

Blocked crossings are penalised rather than forbidden, reflecting that vehicles may cross solid markings in practice, albeit rarely. High-level events are derived as described in Sec.~\ref{sec:grouping}. Additionally, a \texttt{start\_lane} event distinguishes modes that share the same future event sequence but start from different lanes:
 
\begin{minted}[fontsize=\scriptsize,bgcolor=blue!5!white]{prolog}
event(start_lane(L), -1) :- at(Trk, L, 0).
event(lane_change(Trk, Dir), S) :- changed_lane(Trk, S, Dir).
event(intersection(Trk, Dir), S) :- at(Trk, L, S), intersection_lane(L, Dir).
\end{minted}

Each resulting stable model constitutes a distinct trajectory mode, e.g., see the scenario of Fig.~\ref{fig:modes} that are subsequently grouped based on an event fingerprint (Sec  \ref{sec:grouping}).
 
\subsection{Trajectory Generation}
\label{sec:trajectory-generation}
Following the general procedure of Sec.~\ref{sec:traj-gen-general}, each mode group $G_i$ is converted into a continuous trajectory via \emph{centerline construction} and \emph{centerline following}:

\textbf{Centerline Construction.}\quad  Each $M_j \in G_i$ yields an ordered sequence of lane identifiers via \texttt{at/3}, from which lane centerlines are retrieved. To obtain the representative path $\mathcal{C}_i$, lane segments of each model are stitched into a continuous polyline. At junctions, a cubic Bézier bridge (a smooth cubic polynomial curve matched in position and tangent to the adjoining segments) is inserted whose longitudinal reach scales with the lateral offset, producing a smooth $C^1$-continuous transition reflecting natural lane-change geometry. The polylines across all $M_j \in G_i$ are then averaged via arc-length parameterisation---each resampled to a common number of points by fractional arc length and averaged element-wise---yielding $\mathcal{C}_i$ as the geometric mean of the group's paths.

\smallskip

\textbf{Centerline Following.}\quad  Assuming constant speed, the agent advances along $\mathcal{C}_i$ at its observed speed at 10\,Hz. If not already within a snap threshold of $\mathcal{C}_i$, a cubic Hermite transition curve connects the current position and heading to the entry point selected by minimising Eq.~\eqref{eq:entry-cost}, arriving with heading aligned to the centerline tangent. The agent then follows $\mathcal{C}_i$ for the remainder of the horizon, yielding $\mathrm{traj}_i \in \mathbb{R}^{N \times 2}$ with $N = T / \Delta t$, $\Delta t = 0.1$\,s.

\section{Empirical Evaluation with Argoverse~2}\label{sec:eval}
We evaluate the proposed method and its application in autonomous driving along three dimensions: {\small\bfseries(1)}~computational efficiency of the ASP solving pipeline; {\small\bfseries(2)}~geometric sufficiency of the generated lane-path modes; {\small\bfseries(3)}~trajectory forecasting accuracy against established baselines.
All evaluations were conducted on a Apple MacBook Pro (M3) using 23,049 examples from the test split of Argoverse~2\footnote{{\bfseries Argoverse 2} \cite{Argoverse2}., \href{https://www.argoverse.org}{www.argoverse.org}. The Argoverse 2 Motion Forecasting subset is a curated collection of 250,000 scenarios for training and validation. Each scenario is 11 seconds long and contains the 2D, birds-eye-view centroid and heading of each tracked object sampled at 10 Hz.}, which had a vehicle as the focal agent (as per the evaluation scheme in~\cite{Argoverse2}). The following parameters were used throughout: lane change budget $\kappa_{\max} = 2$, minimum recovery distance $\theta_{\mathrm{lc}} = 50$\,m, and a 
6-second prediction horizon at 10\,Hz ($N = 60$ steps).

\subsection{Runtime Complexity}
\label{sec:runtime}
\begin{figure}[t]
    \centering
    \includegraphics[width=\linewidth]{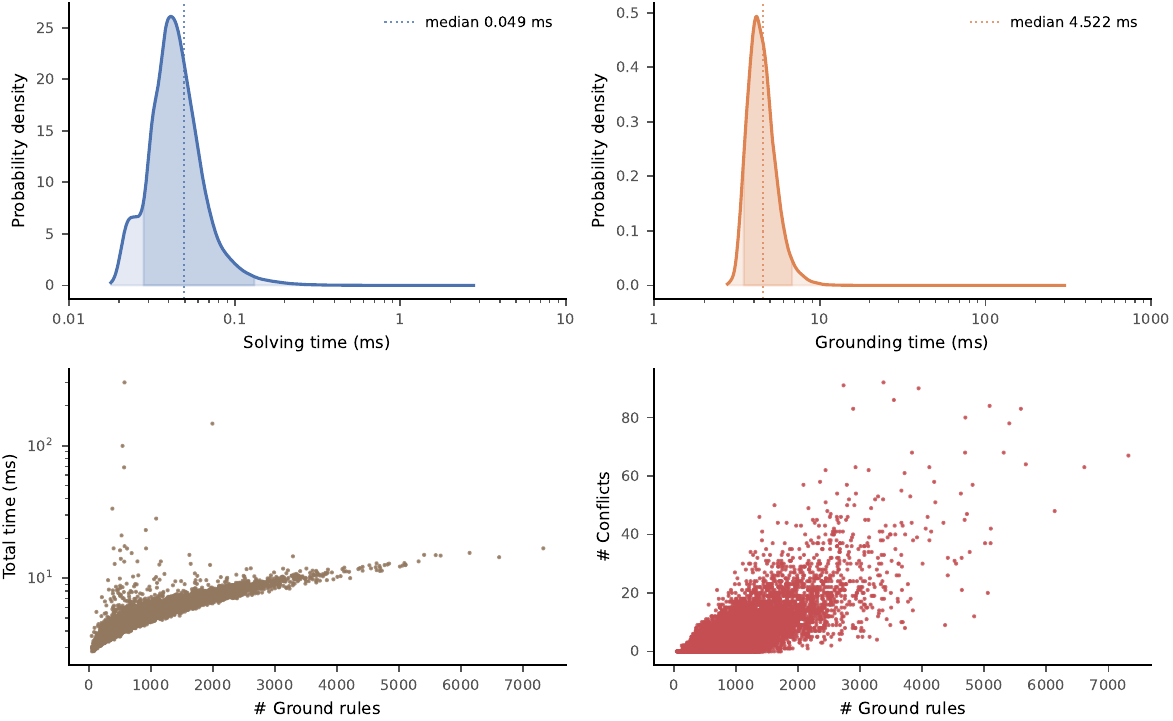}
    \caption{{\bfseries ASP statistics.} \textit{Top:} distributions of solving time (left) and grounding time(right) on a log scale; dashed lines indicate medians. \textit{Bottom:} total runtime (left) and number of conflicts (right) as a function of the number of ground rules.}
    \label{fig:asp_stats}
\end{figure}

Fig.~\ref{fig:asp_stats} reports grounding and solving statistics. Grounding dominates, with a mean of 4.78\,ms (std 2.48\,ms, max 300\,ms). In contrast, solving requires only 0.06\,ms on average. This behaviour is consistent with the modelling strategy of deriving reachability separately from the non-deterministic mode selection, thereby shifting much of the structural computation into rule instantiation and propagation while leaving the solver with a comparatively small combinatorial search space. Additionally, the constant-speed and single-agent assumptions likely contribute to the fast solving speed. Relaxing the constant-speed assumption or extending to multi-agent settings would increase grounding cost, and characterising this scaling behaviour remains future work. The strong correlation between ground rules and total runtime ($\rho = 0.948$) confirms grounding as the primary cost driver, with the right-skewed tail attributable to dense intersection scenarios. Some outliers exhibit disproportionately high runtime relative to their grounding size, likely due to complex local connectivity, such as roundabouts, that increases the number of admissible transitions per step and thus the combinatorial search space, independent of the total lane count. Solver behaviour is lightweight (mean conflicts 3.19, choices 10.24), reflecting effective constraint pruning. On average, 5.88 stable models (std 3.28, min 2, max 28) collapse to 2.94 trajectory modes (std 1.64, min 1, max 14), confirming that event-based grouping substantially reduces the hypothesis space. Scene complexity varies widely (mean 67.10 lanes, std 31.46, max 264), reflecting the diversity of urban layouts in Argoverse~2. More generally, the admissible mode space is shaped by domain-specific constraints such as lane-change spacing and traversal bounds, introducing an explicit tradeoff between behavioural coverage, normative plausibility, and computational tractability.

\subsection{Lane Path Coverage}
We evaluate geometric sufficiency by measuring the fraction of ground truth agent positions across the 6-second horizon that fall within lane polygons of the generated modes. Table~\ref{tab:coverage} reports coverage under three conditions: (i)~\emph{Reachable union}: all lanes across all modes merged, giving an upper bound; (ii)~\emph{Oracle mode}: the single mode with highest coverage, measuring best-case performance; and (iii)~\emph{Ranked mode}: the mode selected by the ASP optimization.
 
\begin{table}[t]
\centering
\caption{\textbf{Lane path coverage}. A 6-second prediction horizon.}
\label{tab:coverage}
\footnotesize\sffamily
\setlength{\tabcolsep}{2pt}
\setlength{\arrayrulewidth}{0.2pt}
\begin{tabular}{
>{\columncolor{lightgray}\bfseries}l
r}
\toprule
\rowcolor{lightgray}
\textbf{Condition} & \textbf{Coverage (\%)} $\uparrow$ \\
\midrule
Reachable union & 90.63 \\
Oracle mode     & 87.78 \\
Ranked mode     & 49.61 \\
\bottomrule
\end{tabular}
\end{table}

\textbf{Results.} \quad The reachable union (90.63\%) and oracle mode (87.78\%) confirm that the lane graph traversal generates a geometrically sufficient hypothesis space and that at least one mode closely tracks the ground truth in most scenarios. The small gap between the two (2.85\,pp) indicates that the best mode rarely requires lanes outside the individually reachable set, i.e., most ground truth paths lie within a single coherent mode rather than requiring a union of several. The ranked mode drops substantially to 49.61\%. This is attributable to the coarse resolution of the preference landscape. The minimisation assigns costs from a small discrete set (0, 1, or 5 per lane change, plus a count penalty), meaning that in many scenarios---particularly those where all reachable paths traverse only dashed markings and involve no lane changes---all stable models receive identical cost. Ties are broken arbitrarily, effectively reducing mode selection to random choice among equally-ranked alternatives. Closing this gap through richer preference models or learned preference weights is a natural extension of the framework. Having established that the symbolic mode space is geometrically well-founded, we evaluate continuous trajectory accuracy.

\begin{table}[t]
\centering
\caption{\textbf{Performance Evaluation}. Results with standard trajectory metrics. Our method is compared with baseline models from~\cite{Argoverse2},
grouped by method type. For scenarios with less than six modes, only the available ones were used.}
\label{tab:trajectory_results}
\scriptsize\sffamily
\setlength{\tabcolsep}{2.5pt}
\renewcommand{\arraystretch}{1.1}
\begin{tabular}{
>{\columncolor{lightgray}\bfseries}l
l
r r r || 
r r r}
\toprule
\textbf{} & \textbf{}
 & \multicolumn{3}{c}{\textbf{top-k=1}}
 & \multicolumn{3}{c}{\textbf{top-k=6}} \\
\cmidrule(lr){3-5}
\cmidrule(lr){6-8}
\rowcolor{lightgray}
\textbf{Model} & \textbf{Type}
 & \textbf{minADE} $\downarrow$
 & \textbf{minFDE} $\downarrow$
 & \textbf{MR} $\downarrow$
 & \textbf{minADE} $\downarrow$
 & \textbf{minFDE} $\downarrow$
 & \textbf{MR} $\downarrow$ \\
\midrule
Constant Velocity \cite{av1}
& \multirow{2}{*}{Heuristic}
& 7.75 & 17.44 & 0.89 & -- & -- & -- \\
Ours (map + const.\ speed)
&
& 6.72 & 16.35 & 0.92 & 4.33 & 10.30 & 0.85 \\
\addlinespace[4pt]
Nearest Neighbor (map) \cite{av1}
& Retrieval
& 6.45 & 15.51 & 0.84 & 4.30 & 10.08 & 0.78 \\
\addlinespace[4pt]
LSTM (map) \cite{av1}
& \multirow{2}{*}{Learning}
& 5.07 & 12.71 & 0.90 & 3.73 & 9.09 & 0.85 \\
WIMP (map + social) \cite{khandelwal2020if}
&
& 3.09 & 7.71 & 0.84 & 1.47 & 2.90 & 0.42 \\
\bottomrule
\end{tabular}
\end{table}

\subsection{Trajectory Evaluation}
\label{sec:traj-eval}
We evaluate single-agent vehicle forecasting using standard trajectory forecasting metrics. These include minimum Average Displacement Error (minADE), minimum Final Displacement Error (minFDE), and Miss Rate (FDE $>$ 2m).

\textbf{Results.} \quad Table~\ref{tab:trajectory_results} compares our results against Argoverse~2 baselines. Against the most directly comparable method---Constant Velocity, which also uses no learned components and shares the constant motion assumption---our method improves top-1 minADE (6.72 vs.\ 7.75), confirming that map-guided mode diversity yields meaningful gains over a single kinematic prediction. Our top-6 performance is comparable to the retrieval-based Nearest Neighbor approach despite using no training data, suggesting that explicit lane-graph reasoning recovers a similar quality of multi-modal diversity to non-parametric data-driven methods. The Miss Rate at top-1 (0.92) is slightly worse than Constant Velocity (0.89), consistent with the ranked-mode coverage gap where the conservative preference signal occasionally selects a mode that is geometrically plausible but behaviourally misaligned with the ground truth. Learning-based methods outperform ours on raw displacement metrics, as expected. WIMP~\cite{khandelwal2020if}, a graph neural network approach, additionally incorporates social context, placing it in a different class entirely.

\begin{figure}[t]
    \centering
    \includegraphics[width=0.9\linewidth]{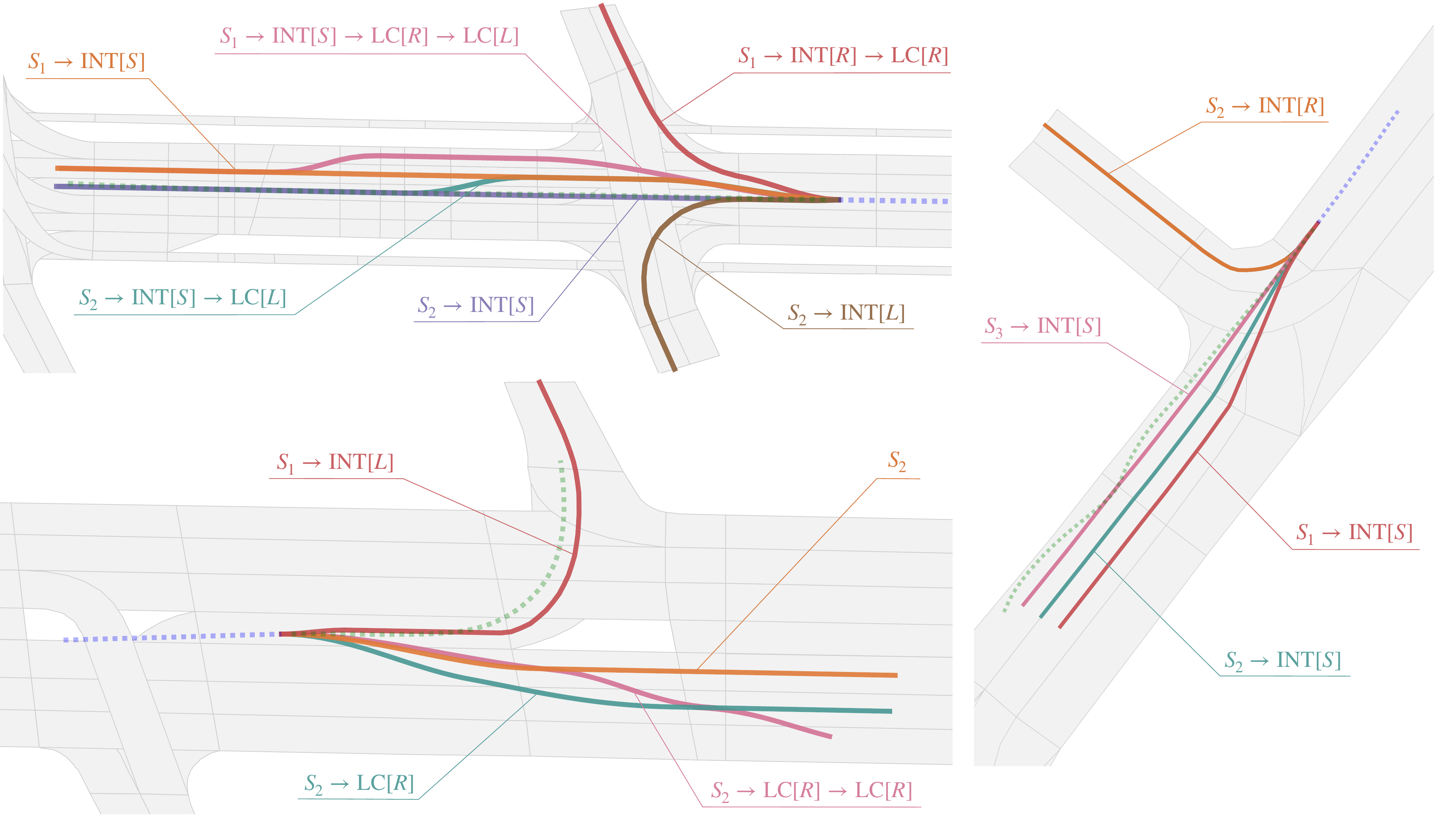}
    \caption{{\bfseries Predicted Trajectories for Three Argoverse~2 Scenarios}. Each coloured path corresponds to one trajectory mode, labelled by its event sequence. $S_n$ denotes the starting node; blue dashed line indicates observation history; green dashed lines represent ground truth. Interpretations for {\sffamily INT[\_] and LC[\_] } are as per Fig. \ref{fig:modes}.}
    \label{fig:out-trajs}
\end{figure}

We note that our method is not developed or positioned as a competitor or replacement of these learned approaches; rather, it constitutes an \emph{explainable baseline} grounded in explicit reasoning over lane-structure and interaction events. Explainability is understood here as \emph{traceability} in that each trajectory is fully traceable to its underlying 
stable model, event sequence, and lane-level path. This property is also illustrated in Fig.~\ref{fig:out-trajs}, which shows a few output examples. As such, the method provides a structured, interpretable foundation complementary to learned predictors, and a natural candidate for hybrid integration as a prior or constraint.

\section{Discussion and Related Work}

Contemporary motion trajectory prediction is dominated by purely quantitative data-driven approaches that learn functional mappings from observed scene contexts to distributions over future trajectories~\cite{shi2025motionsurvey}. Transformer-based architectures have emerged as the dominant paradigm \cite{zhou2023query,zhang2024demo,knoche2025donut}. For example, QCNet~\cite{zhou2023query} reformulates agent--scene interaction through a query-centric representation that achieves roto-translation invariance and supports streaming scene encoding, attaining state-of-the-art performance on Argoverse~2. DeMo~\cite{zhang2024demo} decouples temporal and social feature extraction into separate pathways, improving efficiency and performance over end-to-end baselines while remaining within the learned latent paradigm. DONUT~\cite{knoche2025donut} pushes further by adopting a decoder-only autoregressive architecture inspired by language modelling, generating trajectory tokens in a single forward pass without an explicit encoder--decoder separation. While these methods achieve strong geometric accuracy on benchmarks, a structural limitation is common to all: the representations that mediate prediction---learned queries, latent modes, decoder hidden states---typically lack explicit interpretable semantics. A trajectory generated by QCNet cannot be explained in terms of which lane the vehicle was predicted to occupy, which intersection manoeuvre was anticipated, or which normative constraint was satisfied. The same geometric output could, in principle, arise from multiple qualitatively distinct behaviours without distinction. DeMo's decoupled features and DONUT's token sequence likewise provide no symbolic account of the behavioural mode selected. This opacity is not a limitation of any individual architecture but an inherent consequence of the end-to-end learning paradigm, where interpretability is not an inherent design objective, and mechanisms for systematic verification or retrospective diagnosability not directly achievable.

\smallskip

Goal-based and hierarchical methods \cite{Goal-LBP2024,Mangalam2021,GANet} address the qualitative vs. quantitative gap by abstracting prediction into geometric targets or waypoints, but these targets are typically learned rather than derived from explicit commonsense reasoning over the environment structure. We take a complementary stance, closer in spirit to physics- and 
rule-based approaches~\cite{shi2025motionsurvey}. Rather than learning to approximate the distribution of future trajectories, we enumerate qualitatively distinct behavioural hypotheses, e.g., pertaining road network geometry. Each generated trajectory is paired with a verifiable event sequence and a lane-level path, making the reasoning process fully traceable. However, the cost is geometric accuracy: a constant-speed kinematic model cannot match the displacement errors of learned predictors trained on hundreds of thousands of scenarios. On the other hand, the benefit is transparency in the sense that each output can be inspected, explained, and verified against the map and domain norms.

\section{Conclusion and Outlook}
We have developed a declarative method for hybrid qualitative-quantitative computation of reasonable trajectories of moving objects. The method is systematically formalised and evaluated with Argoverse 2, a state-of-the-art, community established benchmark in the domain of autonomous driving. The work is a departure from monolithic end-to-end approaches where explainability and interpretability are not design considerations per se. Our outlook is geared towards application in integrated perception and control, by combining motion trajectory commonsense about moving objects with (ASP-based) high-level visual sensemaking capabilities of \emph{neurosymbolic vision} techniques (e.g., \cite{kr2025-asp-visualcomm,AIJ2021-Suchan,probRanking-Monsen-2025}, as well as with emerging Vision–Language Models (VLMs; e.g., \cite{vlm-foiks-2026,Simeoni2025_dinov3}) for low-level visual feature analysis in real-time vision tasks.

\medskip

\textbf{Acknowledgements}
We acknowledge funding by the Swedish Research Council (Vetenskapsrådet -- VR) through the project \emph{Counterfactual Commonsense} (Project ID: 2022-02960\_VR).

\smallskip

\textbf{Disclosure of Interests.}
The authors have no competing interests to declare that are relevant to the content of this article.

\bibliographystyle{splncs04}

\end{document}